%% file: cce_heterogenous.tex
\documentclass[a4paper]{article}

\usepackage{INTERSPEECH2018}
\usepackage{bm}
\usepackage{amsmath, amssymb}
\usepackage{algorithm, algorithmic}
\usepackage{todonotes}

\title{Contextual Slot Carryover for Disparate Schemas}
\name{Chetan Naik, Arpit Gupta, Hancheng Ge, Lambert Mathias, Ruhi Sarikaya}
\address{Amazon Alexa Machine Learning}
\email{\{chetnaik, arpgup, ghanche, mathiasl, rsarikay\}@amazon.com}
\begin{document}

\maketitle
\begin{abstract}
 \input{abstract}

\end{abstract}
\noindent\textbf{Index Terms}: {spoken dialog, state tracking, anaphora resolution, contextual understanding.}

\section{Introduction}
\input{introduction}

\section{Approach}
\label{sec:approach}

\subsection{Task Definition}
\label{sec:task}
\input{task_definition}

\subsection{Model Definition}
\label{ssec:model}
\input{model_definition}

\section{Experiments}
\input{experiments}
\label{sec:experiments}


\section{Conclusion}
\input{conclusion}
\label{sec:conclusion}

\bibliographystyle{IEEEtran}

\bibliography{mybib}

\end{document}

%% file: abstract.tex
In the slot-filling paradigm,  where a user can refer back to slots in the context during a conversation, the goal of the contextual understanding system is to resolve the referring expressions to the appropriate slots in the context. In large-scale multi-domain systems, this presents two challenges - scaling to a very large and potentially unbounded set of slot values, and dealing with diverse schemas. We present a neural network architecture that addresses the slot value scalability challenge by reformulating the contextual interpretation as a decision to  {\it carryover} a slot from a set of possible candidates. To deal with heterogenous schemas, we introduce a simple data-driven method for transforming the candidate slots. Our experiments show that our approach can scale to multiple domains and provides competitive results over a strong baseline.

%% file: introduction.tex
Slot-filling based spoken language understanding (SLU) system is often a central component~\cite{Mesnil2013InvestigationOR} in conversational systems. A major challenge in the slot-filling paradigm is to handle conversational context, where a user utterance can refer back to a set of slots implicitly or explicitly.

Traditionally, the contextual interpretation of slots has been cast as a coreference resolution problem. There is a rich body of work on coreference resolution for written text, which rely on clustering\cite{wellner2003towards, stoyanov2012easy, culotta2007first} or ranking mention pairs~\cite{durrett2013easy,wiseman2015learning}. These have been extended to spoken dialog, by adding discourse specific features~\cite{rao2015dialogue, strube2003machine, Stent2010InteractionBD, liu2015deep, Williams2013TheDS, eckert1999resolving}. However, most of these approaches follow a pipelined model of mention detection followed by coreference resolution; where linguistic features, syntax and discourse features are usually applied. In contrast, our proposed formulation does not rely on explicit linguistic features such as gender and type agreement, which are hard to acquire across languages. Furthermore, we can generalize the solution to sub-tasks such as zero pronouns naturally, as we don't have to explicitly identify the anaphoric mentions.

Another challenge is dealing with multi-domain language understanding systems, where each domain has its own schema to represent slots and intents. Domains are developed mostly independently and hence do not share a common schema. Also, dialog assistants are now being extended by community developers using services such as Google DialogFlow\footnote{https://dialogflow.com/API.AI} or Alexa Skills Kit \cite{justASK}. Domains developed by external developers are completely outside of a central repository of domain and slots, hence no assumption can be made about their schema. The lack of shared schema makes it hard to maintain contextual slots across domain boundaries. Table ~\ref{tbl:cca} shows an example conversation a user may have with such a dialog system. In this example user interacts with three domain: Weather, LocalSearch and Traffic. Each domain has its own schema to represents slots.  The user starts off by asking weather in a city, follows it up with question about restaurant that serves \textit{mexican} cuisine, finally she asks about directions to the restaurant. In this example, we showcase the challenges of multi-domain system. Weather and LocalSearch use different schema to store information about the location, using slot keys {\it WeatherLocation} and {\it City} respectively. For carrying conversation across domains from \textbf{U1} to \textbf{U2}, we need to be able to transform the slot [{\it WeatherLocation: san francisco}] to [{\it City: san francisco}], without having access to a common schema. Even within a domain there could be issues of diverse schemas -  \textbf{U2} and \textbf{V2}, the domain chooses to have different schemas to represent its user and system turns.  To make the task more complex, some domains choose to represent all its slots as just  a generic label {\textit{Entity}}.

\newcommand{\light}{\selectfont}

\begin{table*}
\begin{center}
\begin{tabular}{|l|l|c|c|}\hline
\multicolumn{1}{|c|}{Domain} & \multicolumn{1}{|c|}{Turns} & \multicolumn{1}{|c|}{Current Turn Slots} & \multicolumn{1}{|c|}{Carried Slots}\\\hline
Weather & {\bf U1}: weather in san francisco & {\light WeatherLocation: san francisco} &\\\hline
Weather & {\bf V1}: weather is rainy and temperature 42F & {\light Temperature: 42F} &\\\hline
LocalSearch & {\bf U2}: any mexican restaurants nearby & {\light PlaceType: mexican restaurants} & {\light City: san francisco}\\\hline
LocalSearch & {\bf V2}: la taqueria is a mile away & {\light Entity: la taqueria} &\\\hline
Traffic & {\bf U3}: thanks, send directions to my phone & &{\light Place: la taqueria} \\
 &  & & {\light Town: san francisco} \\\hline
\end{tabular}
\end{center}
\caption{Heterogenous schema: An example multi-domain dialog. The slot (WeatherLocation, san francisco) in in the Weather domain (U1) when carried over to LocalSearch domain (U2) is mapped to the slot (Town, san francisco).}
\label{tbl:cca}
\end{table*}

There has been work on improving semantic frame error rate for current turn by leveraging context turns by encoding dialog states.~\cite{bapna2017sequential} compare various approaches encoding context, ~\cite{chen2016end} describe a memory network architecture for knowledge carryover,  ~\cite{yann2014zero} add semantic context from the frame, ~\cite{xu2014contextual} use context features for domain classification. Our work differs from this body of literature; we keep the system for semantic frame prediction fixed and explore methods to explicitly add slots from previous turn. While the previous work assumes existence of a dialog manager which can be used to keep track of entities from previous turns, we make no such assumptions.

A closely related task is dialog state tracking~\cite{Williams2013TheDS, hori2016dialog, Henderson2013DeepNN}, where the system has to predict a set of slot-value pairs which matches the contents of the current segment. Usually, state trackers produce a distribution over all possible slot-value pairs; this does not scale for open-ended slot values (such as Date or Time), as well as slots whose values are constantly being updated (such as Songs, Movies). Our approach avoids this by reformulating the tracking problem as a {\it carryover} action for the current turn. More closely related is  frame-tracking \cite{frametracking2017}, which was introduced as an extension to state tracking, here the slots need to be tracked over multiple frames and maintain reference to original frame. A key difference here is that our formulation deals with the issue of disparate labels over a large-scale multi-domain system.

In this paper, we present a neural network architecture that addresses the challenges above. Main contributions of the paper.
\begin{enumerate}
\item We present the task of tracking slots in a conversation as a \textit{carrryover} decision. This allows us to scale to a potentially unbounded set of slot values, and allows us to generalize anaphora resolution to both explicit and implicit references. 
\item We address diverse schema challenge by leveraging label embeddings (see sec ~\ref{cand_gen}) to generate potential candidates to be carried.
\item We show our proposed model outperforms a strong rule-based baseline. We also demonstrate via experiments why our task is more complex than dialog state tracking by benchmarking our approach on DSTC-2 as well as on a dataset collected from a real virtual assistant device. 
\end{enumerate}

%% file: task_definition.tex
We define a a dialog turn at time $t$ as the tuple $\{a_t, \bm{S}_t, \bm{w}_t\}$, where $\bm{w_t} \in \mathcal{W}$ is  a sequence of words $\{w_{it}\}_{i=1}^{N_t}$; $a_t \in \mathcal{A}$ is the dialog act; and $\bm{S}_t$ is a set of slots, where each slot $s$ is a key value pair $s=\{k,v\}$, with $k\in \mathcal{K}$ being the slot name (or slot key), and $v\in \mathcal{V}$ being the slot value. $\bm{u}_t=\{a_t^u, \bm{S}_t^u, \bm{w}_t^u\}$ represents a user-initiated turn and $\bm{v}_t=\{a_t^v, \bm{S}_t^v, \bm{w}_t^v\}$ represents a system initiated turn.
 
Given a sequence of $D$ user turns $\{\bm{u}_{t-D+1}, \dots, \bm{u}_{t-2}, \bm{u}_{t-1}\}$; and their associated system turns $\{\bm{v}_{t-D+1}, \dots, \bm{v}_{t-2}, \bm{v}_{t-1}\}$\footnote{For simplicity we assume a turn taking model - a user turn and system turn alternate.}; and the current user turn $\bm{u}_t$, the task is to predict a carryover decision over each of the candidate slots in $C(\bm{S})=\bigcup\limits_{i \in {u,v}, j=t-D+1}^{t-1} \bm{S}^i_j$ i.e  we carryover slot $s \in C(\bm{S})$ to turn $u_t$ if  $P(+1|s, \bm{u}_t, \bm{u}_{t-D+1}^{t-1}, \bm{v}_{t-D+1}^{t-1}) > \tau$, where $\tau$ is a decision threshold to be optimized. This formulation allows us to scale to potentially unbounded slots and also handle diverse schemas as we will discuss later.

%% file: model_definition.tex
\begin{figure*}[ht]
\begin{center}
\includegraphics[width=\textwidth]{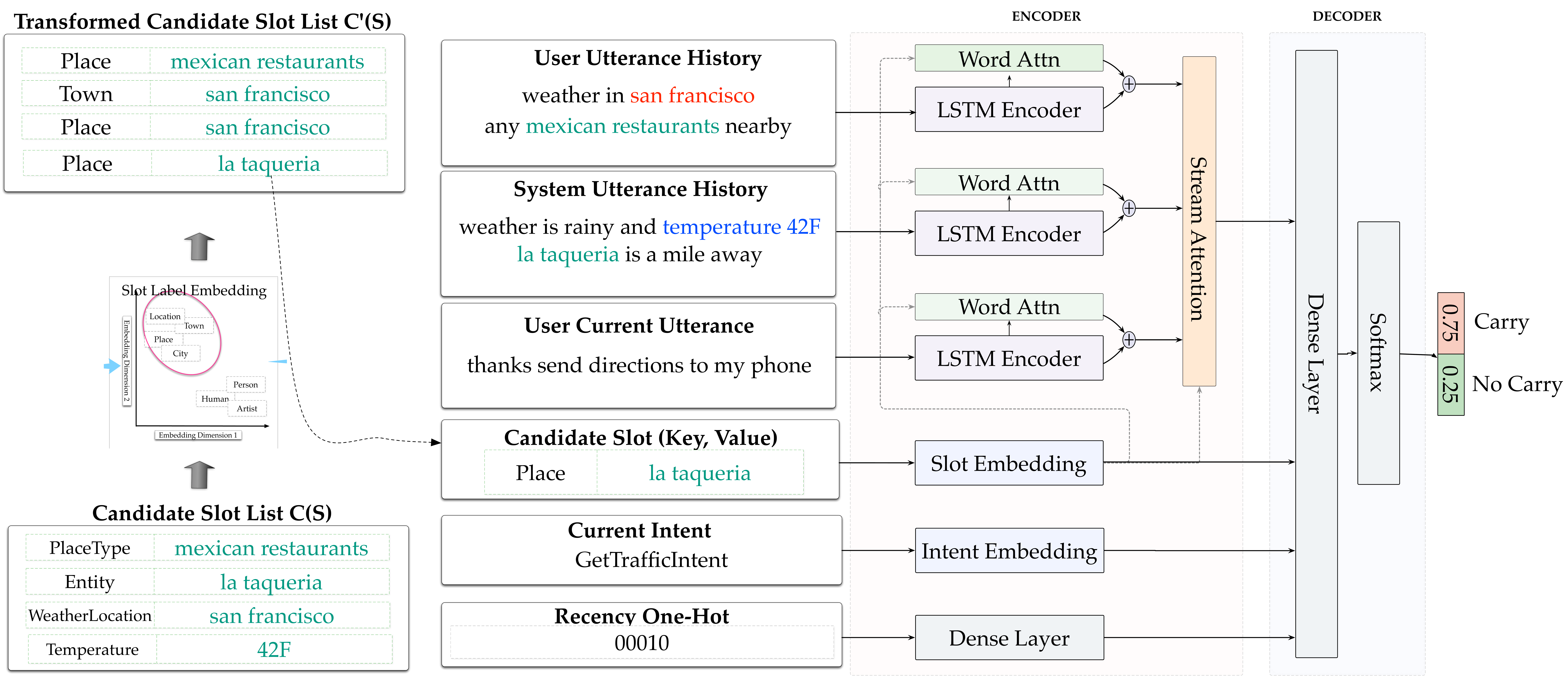}
\caption{Encoder-decoder architecture for slot carryover model. First candidate generation is run using the slot embeddings as described in Section~\ref{cand_gen}. Then for each slot in the transformed candidate list the model makes an independent decision for carryover. For the example dialog, and the candidate slot  [PlaceName: la taqueria], fixed length encodings - over candidate slot features, the dialog history for both user and system, and the current turn - are used to make the carryover decision at the current turn.}
\label{fig:model}
\end{center}
\end{figure*}

We use an encoder-decoder approach as shown in Figure~\ref{fig:model} to classify each candidate slot as being relevant for the current turn.
into

\subsubsection{Candidate Slot Generation}
\label{cand_gen}
As shown in Table~\ref{tbl:cca}, the schemas associated with each turn can be in completely different label spaces. So, we use slot key embeddings to map the keys of the candidate slots in $C(\bm{S})$ into the schema associated with the current domain.  We use pre-trained word embeddings as the source for computing the slot key embeddings. 

For each slot name $k$ we compute its label embedding by averaging over the $M_k$ associated slot value embeddings. For multi-word slot values, the embedding is constructed by averaging the associated word embeddings. 
\begin{eqnarray}
\label{eq:v}
\Phi_W(v) = \frac{1}{|w \in v|}\sum_{w \in v} \Phi_W(w)\\
\label{eq:k}
\Phi_K(k) = \frac{1}{M_k}\ \sum_{i=1}^{M_k} \Phi_W(v_i)
\end{eqnarray}

We now construct the transformed candidate set as $C'(\bm{S}) = \{ (k',v) |  (k,v) \in C(\bm{S})\ \& \  \Phi_K(k) \odot \Phi_K(k') > \beta\}$, where, $\odot$ is the dot product and $\beta$ is a tunable threshold over the development set.

\subsubsection{Dialog Encoding}
\label{ssec:stream}
We first embed the words in the utterance sequence $\bm{w}$ using word embeddings $\Phi_W$ \cite{pennington2014glove}, to get the sequence ${\bm x}$, which are then fed into an LSTM to recursively encode the current turn, the turns in context associated with the user and the system respectively:
\begin{eqnarray}
 &{\bm h}_{t}^u = LSTM({\bm x}_t^u)\\
 &{\bm h}_{t-1}^u = LSTM({\bm x}_j^u),\ t-D+1 \leq j < t\\
  &{\bm h}_{t-1}^v = LSTM({\bm x}_j^v),\ t-D+1 \leq j < t
\end{eqnarray}

%
%

The LSTM is stateful i.e the  LSTM output of the last token in the utterance is fed as an initial state to the LSTM for the next utterance. 

 \subsubsection{Encoding Dialog Act}
 The dialog act for the current turn $a_t$ is encoded into a fixed length vector of dimension $D_A$, using the intent embedding dictionary $\phi_A$ as :

For each dialog act $k$ we compute its embedding by averaging over the $M_k$ associated utterance embeddings. For each utterance we calculate its embedding is constructed by averaging the associated word embeddings. 
\begin{eqnarray}
\label{eq:u_i}
\Phi_U(u) = \frac{1}{|w \in u|}\sum_{w \in u} \Phi_W(w)\\
\label{eq:a_i}
\Phi_A(k) = \frac{1}{M_k}\ \sum_{i=1}^{M_k} \Phi_U(u_i)
\end{eqnarray}

  
 \subsubsection{Encoding Candidate Slot}
 The candidate slot $s=(k,v)$ is encoded into a fixed length vector $D_s$ as a concatenation of the slot key embedding and the slot value embedding. 
\begin{equation}
h_{s}=\phi_K(k) \oplus \phi_W(v)
\end{equation}
 
Here, $\oplus$ implies concatenation; $\phi_W(v)$ and $\phi_K(k)$ is as defined in Equation~\ref{eq:v} and Equation~\ref{eq:k}  respectively. 
 
\subsubsection{Recency Encoding}
The slot distance $d_s$, defined as the integer offset of the candidate slot from the current turn, is encoded as  one-hot $\{0,1\}^{|D|}$. The final distance encoding vector can then be constructed using an affine transform:
\begin{equation}
h_{d} = W_d * OneHot(d_s) + b_d
\end{equation}

 \subsubsection{Attention Mechanism}
 We consider two levels of attention - the word level attention allows the model to focus on individual mentions in the utterance that influence the slot carryover decision, and the stream level attention which allows the model to focus on specific streams (user and system) in the dialog. 
 
\textbf{Word Attention}: For each stream defined in Section~\ref{ssec:stream}, we attend over the words in that stream and compute a per-stream context vector. For stream vector sequence ${\bm h}_{t}^u$, and slot embedding $h_s$, we compute the word level attentional context vector as:
\begin{eqnarray}
& e_{js} = g(h_{tj}^u, h_s)\\
& \alpha_{js} = \text{softmax}(e_{js})\\
& c_t^u=\sum_{j=1}^{N_t} \alpha_{js} h_{tj}^u
 \end{eqnarray}
 Here $j$ an index into ${\bm h}_{t}^u$, represents the hidden encoding of the associated input word at that position. We then compute the importance of the word to the slot as the similarity defined in $e_{js}$; obtain the normalized weights $\alpha_{js}$ that is then used to compute the weighted context vector $c_t^u$. Similarly, we can compute $c_{t-1}^v$ and $c_{t-1}^u$.
 
\textbf{Stream Attention}: As before, we attend over each individual stream $c_k \in \{ c_{t-1}^v, c_{t-1}^u, c_{t}^u \}$ to obtain the final context vector $h_c$\footnote{If word attention is turned off we choose the final state from each stream LSTM to construct $h_c=h_{t,final}^u + h_{t-1,final}^v + h_{t-1,final}^u$}
 \begin{eqnarray}
& e_{ks} = g(c_{k}, h_s)\\
& \alpha_{ks} = \text{softmax}(e_{ks})\\
& h_c=\sum_{j=1}^{3} \alpha_{ks} c_{k}
 \end{eqnarray}

%
%
%
%

\subsubsection{Decoder}
The vectors from the encoders are concatenated and sent to the final softmax layer to get the class probabilities as follows
\begin{eqnarray}
&z = h_c \odot h_a \odot h_s \odot h_d\\
&\hat{o} = \text{softmax}(W_{decoder} * z + b_{decoder})
\end{eqnarray}

%% file: experiments.tex
\subsection{Data Setup}
For the experiments, we use subset of data collected on a commercial voice assistant.
Table~\ref{tbl:data} summarizes the statistics in the training, development and test sets across different domains. Around 20\% of the sessions have utterances from multiple schema. Also, as expected for a voice assistant, we have a significant imbalance where the number of positive candidate slots are much smaller than number of possible candidates for each turn. This is due to cross domain interactions which follow each other but are not part of the same goal, which is common in a digital assistant. Furthermore, some domains chose not to associate any label with an entity mention which we represent as \textit{\textbf{Entity}} slot this results in very large number of potential candidates as we consider all possible target slots in the current domain for such entities.

To demonstrate complexity of our data we also report results on dataset released in Dialog State Tracking Challenge ~\cite{henderson2014second}. We modified the dataset  to fit our caryryover task accordingly. We consider only the 1-best ASR and 1-best SLU hypothesis. Unlike our commercial dataset, in DSTC the slots tracked as part of the goal only occur from the user turn. So, we remove candidates from system turns as a pre-processing step. Also, in DSTC task dialogs can be system initiated but for our task is always user initiated, hence we remove the first system turn.

\subsection{Training Setup and Evaluation Metrics}
We introduce two baselines. The `Naive Baseline' system carries over all the slots from the most recent turn in the dialogue session.This is because, most recent entities are more likely to be referred to by the users in a spoken dialogue system.  We also use a  stronger elaborate `Rule Baseline', where we detect the referring expression, and for each referring expression, linguistic and semantic features are used to retain only those antecedent candidate slots that agree in gender, number and type. Algorithm~\ref{alg:rule_loc} shows an example rule that executes for the use case U2 in Table~\ref{tbl:cca}. 

\begin{table}[ht]
\begin{center}
\resizebox{0.5\textwidth}{!}{
\begin{tabular}{|c||c|c|c|} \hline
Domain & Train & Dev & Test  \\\hline
Music & 4587 & 558  & 580 \\
Weather & 6067 & 738 & 729 \\
Local Businesses & 1439  & 162 & 185 \\
Video & 1000 & 99  & 141 \\
Q\&A & 386  & 49 & 48 \\
Home Automation & 1945 & 230 & 291  \\
Others & 1481 & 163 & 178  \\\hline
Total & 16905 & 1999 & 2152  \\\hline
Avg. turns per session & 2.2 & 2.14 & 2.18 \\\hline
\%age of disparate schema sessions & 19.86 & 18.75 & 20.53  \\\hline
Avg. positive carryover candidates per turn & 0.37 & 0.39 & 0.35 \\\hline
Avg. negative carryover candidates per turn & 4.07 & 4.05 & 4.00 \\\hline
\end{tabular}}
\end{center}
\caption{Contextual Carryover Data Setup}
\label{tbl:data}
\vspace{-7mm}
\end{table}

\begin{algorithm}
\caption{Example rule carrying over City slot as it is compatible in type to anaphor "there"}
\label{alg:rule_loc}
\begin{algorithmic}
\IF {$ReferringPhrase.Type == CandidateSlot.Type$}
\IF {$CandidateSlot.Type == City$}
\IF {$ReferringPhrase.Value == "there"$}
\STATE $CarriedSlots += CandidateSlot$
\ENDIF
\ENDIF
\ENDIF
\end{algorithmic}
\end{algorithm}

For the model, we initialize the word embeddings using 300 dimensional pre-trained GloVe~\cite{Pennington2014GloveGV} vectors. The model is trained using mini-batch SGD with Adam optimizer~\cite{Kingma2014AdamAM} with standard parameters to minimize the class weighted cross-entropy loss. 
In our experiments, we use 128 dimensions for the LSTM hidden states and 256 dimensions for the hidden state in the decoder. Similar to~\cite{Wiseman2015LearningAA}, we pre-train a LSTM for the named entity recognition task and use this model to initialize the parameters of the LSTM based encoders. All model setups are trained for 20 epochs with early stopping criterion optimised on a dev set.
We only select those slots as the final hypothesis, whose $\tau > 0.5$, which was optimized over the dev set. For each utterance, independent carryover decisions are made for each candidate slot. We evaluate the models by comparing the hypothesis and reference slots to measure precision, recall and F1 scores.

\subsection{Results and Discussion}
Table~\ref{tbl:results} shows the cumulative impact of various training strategies of our proposed model. We see that compared to a strong rule-based system, our proposed approach gives significant gains in accuracy. Our proposed encoder-decoder improves upon the strong rule based baseline. Adding word attention helps improve the precision and F1 but at the cost of recall; the results are significant compared to the system without attention. Intuitively, the slot value matching referring tokens in the dialog turn indicates that it is relevant to the conversation. The word attention model captures this intuition as part of the model learning process. This alleviates the need to explicitly define semantic  type similarity features, and detecting anaphoric mentions like we do in the rule based system. Adding stream attention improves recall, but the overall F1 degrades. Stream attention, which helps isolate user and system turns did not help; we speculate that the distance feature already captures this, and there is insufficient data to train this appropriately.

For completeness we also include performance on the public DSTC2 dataset in Table~\ref{tbl:dstc_results} . We do not claim to be solving DSTC2 but only use this dataset as a comparison of task complexity - the DSTC2 task is relatively simple as evidenced by the naive baseline having a high F1 score on this task, but very low on our commercial assistant task.

\begin{table}[t]
\centering
\small
\begin{tabular}{|l||c|c|c|} \hline
Method & Precision & Recall & F1\\\hline\hline
Naive Baseline & 17.01 & 92.50 & 28.74 \\
Rule Baseline & 91.79 & 67.11 & 77.53 \\
Encoder-Decoder & 73.31 & 96.17 & 83.20 \\
+ word attention & \textbf{75.76} & 94.65 & \textbf{84.16} \\
+ stream attention & 73.48 & \textbf{96.18} & 83.31 \\\hline
\end{tabular}
\caption{Modeling Architecture Impact on Accuracy for Multi-Domain Dataset}
\label{tbl:results}
\vspace{-5mm}
\end{table}

\begin{table}[t]
\centering
\small
\begin{tabular}{|l||c|c|c|} \hline
Method & Precision & Recall & F1\\\hline\hline
Naive Baseline & 80.58 & 75.44 & 77.93 \\
Encoder-Decoder & 97.22 & 95.30 & 96.24 \\
+ word attention & 97.20 & 97.65 & 97.42 \\
+ stream attention & 97.23 & 95.61 & 96.42  \\\hline
\end{tabular}
\caption{Results on DSTC2 dataset}
\label{tbl:dstc_results}
\vspace{-3mm}
\end{table}

\begin{table}[h]
\centering
\small
\begin{tabular}{|l||c|c|c|} \hline
Category & Precision & Recall & F1\\\hline\hline
Within-Domain & 78.19 & 95.7 & 86.09 \\
Cross-Domain & 61.66 & 87.05 & 72.19 \\\hline
\end{tabular}
\caption{Cross-domain accuracy of encoder-decode w/ word attention for the model in Table~\ref{tbl:results}}
\label{tbl:results_domain}
\end{table}

%% file: conclusion.tex
In this work, we presented the task of contextual carryover of slots in a multi-domain large-scale  dialog system. To address the scalability of the solution over a large set of slot values we re-formulated this as a slot carryover decision to identify the most relevant set of slots at the current turn. Furthermore, we proposed an efficient way to leverage label embeddings to deal with heterogeneous schemas. We presented empirical results demonstrating the efficacy of our neural network formulation over a strong rule-based baseline. We also quantified the gains from various components of the proposed approach. 
